\documentclass{llncs}
\usepackage{graphicx}
\usepackage{amsmath,amssymb} 
\usepackage{color}
\usepackage{booktabs} 
\usepackage[width=122mm,left=12mm,paperwidth=146mm,height=193mm,top=12mm,paperheight=217mm]{geometry}

\usepackage{setspace}

\begin{document}

\setlength{\abovecaptionskip}{1mm}
\setlength{\belowcaptionskip}{1mm}

\pagestyle{headings}
\mainmatter
\def\ECCV18SubNumber{2704}  

\title{PM-GANs: Discriminative Representation Learning for Action Recognition Using Partial-modalities} 


\author{Lan Wang$^1$, Chenqiang Gao$^1$, Luyu Yang$^1$, Yue Zhao$^1$, Wangmeng Zuo$^2$, \\and Deyu Meng$^3$ }

\institute{$^1$Chongqing University of Posts and Telecommunications, Chongqing, China\\$^2$Harbin Institute of Technology, Harbin, China\\
$^3$ Xi'an Jiaotong University, Xi'an, China}

\maketitle

\begin{abstract}

Data of different modalities generally convey complimentary but heterogeneous information, and a more discriminative representation is often preferred by combining multiple data modalities like the RGB and infrared features. However in reality, obtaining both data channels is challenging due to many limitations. For example, the RGB surveillance cameras are often restricted from private spaces, which is in conflict with the need of abnormal activity detection for personal security. As a result, using partial data channels to build a full representation of multi-modalities is clearly desired. In this paper, we propose a novel Partial-modal Generative Adversarial Networks (PM-GANs) that learns a full-modal representation using data from only partial modalities. The full representation is achieved by a generated representation in place of the missing data channel. Extensive experiments are conducted to verify the performance of our proposed method on action recognition, compared with four state-of-the-art methods. Meanwhile, a new Infrared-Visible Dataset for action recognition is introduced, and will be the first publicly available action dataset that contains paired infrared and visible spectrum.

\end{abstract}

\section{Introduction}

Human action recognition   \cite{varol2017long,fernando2017rank,li2018videolstm,simonyan2014two,wang2015action,yang2014novel} aims to recognize the ongoing action from a video clip. As one of the most important tasks in computer vision, action recognition plays a significant role in many useful applications like video surveillance  \cite{tran2015learning,karpathy2014large}, human-computer interaction \cite{rautaray2015vision,lindtner2014emerging} and content retrieval \cite{bouwmans2014robust,yang2014exploiting}, with great potentials in artificial intelligence. As a result, massive attention has been dedicated to this area which made large progress over the past decades. Most state-of-the-art methods have contributed to the tasks in visible imaging videos, and shows saturated performances among the widely-used benchmark datasets including KTH \cite{veeriah2015differential} and UCF101 \cite{van2015apt}. Generally speaking, the task of action recognition is quite well-addressed and has already been applied to real-world problems.

\begin{figure}[htbp]
	\centering
	\includegraphics[width=9cm]{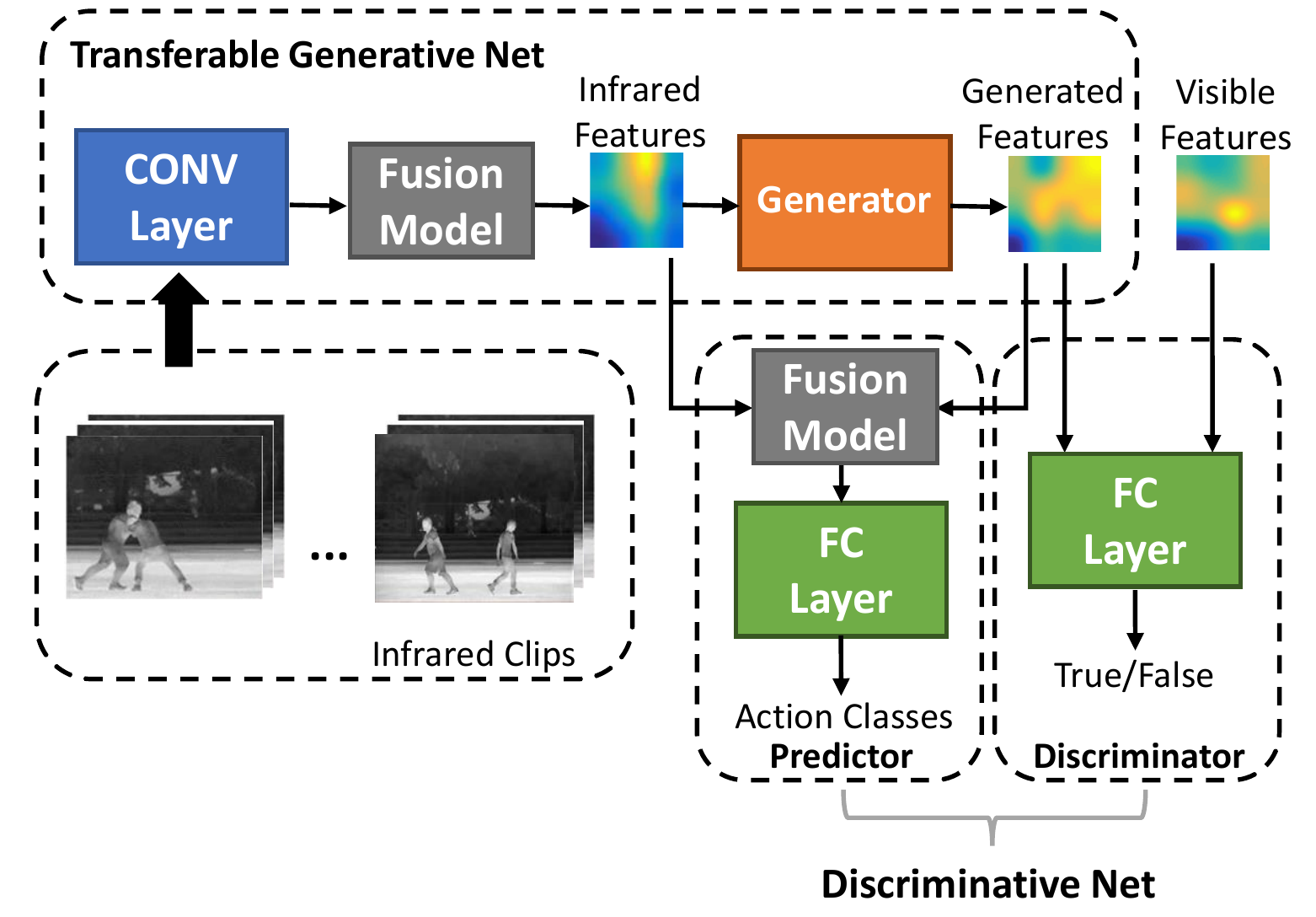}
	\vspace{-1ex}
	\caption{The framework of the proposed Partial-modal Generative Adversarial networks (PM-GANs).
		Infrared video clips are sent to the transferable generative net to produce fake feature representation of the visible spectrum. And the discriminator attempts to distinguish between the generated features and the real ones. The predictor construct a full representation using the generated features and infrared features to conduct classification.}
	\label{fig:framework}
	\vspace{-0.6cm}
\end{figure}

However, there are still many occasions where visible imaging is limited. First, the RGB cameras rely heavily on the light conditions, and perform poorly when light is insufficient or over-abundant. Action recognition from night-view RGB data remains a rather difficult task. Moreover, as an act to protect the fundamental human dignity--Privacy, RGB cameras are strictly restricted from most private areas including the personal residential, public washroom where abnormal human activities are likely to threat personal security. Infrared cameras, that captures the heat radiation of objects, are excellent alternatives in these occasions \cite{gao2016infar}. The application of thermal imaging in military affairs and police surveillance has continued for years, and has more potentials beyond the government use. With many advantages over the RGB camera, it is predicted that infrared cameras will become more common in public spaces like hospitals, nursing centers for elderly and home security systems \cite{liu2018transferable}.

While infrared cameras can fill the limited spots of RGB cameras, many visible features are nevertheless lost in the infrared spectrum due to their similarity in temperature \cite{zollhofer2014real,wu2017rgb}. Visible features like color, texture are effective clues in activity representations. Since the two are complementary to each other, it is desired to utilize both visible and infrared features to benefit the task of action recognition. Furthermore, it will be more desired to utilize both feature domains when ONLY infrared data is available. In the previous cases when the demand of abnormal action recognition and the demand of privacy conflicted, it will be great if we can obtain both infrared and visible features, while use only the infrared data. The question is, how can one obtain visible features when the visible data is missing? The situation is not unique to the task of action recognition. 
In fact, data with different modalities of complementary benefits widely exist in multimedia such as systems with multiple sensors, product details with combined information of text description and images \cite{pereira2014role}. Here we are inspired by the intra-modal feature representations to make up for the missing data using adversarial learning with the available part of the data channel.

Recently, much attention has been given to cross-modal feature representations \cite{wei2017cross,kang2015learning,feng2014cross,castrejon2016learning} dealing with unpaired data, which maps multiple feature spaces onto a common one, or to generate a different representation via adversarial training. 
The basic model of generative adversarial networks (GANs) \cite{goodfellow2014generative,denton2015deep,radford2015unsupervised} consists of a generative model $G$ and a discriminative model $D$. 
Many interesting image-to-image translations such as genre translation, face and pose transformation indicate the broader potentials of GANs to explore the hidden correlations in cross-modal representations \cite{liu2018transferable,peng2017cm}. Inspired by this, we therefore seek an algorithm that can translate from the infrared representation to the visible domain, which allows us to further exploit the benefits of both feature spaces with only part of the data modalities. More generally speaking, we aim at an architecture that learns a full representation for data of different modalities, using partial modalities. Different from the existing works of cross-modal which seeks a common representation from different data spaces, our goal is to exploit the transferable ability among different modalities, which is further utilized to construct a full-modal representation when only partial data modalities are available.

With a completely different target, in this paper we propose a novel partial-modal Generative Adversarial Networks (PM-GANs), which aims to learn the transferable representation among data of heterogeneous modalities using cross-modal adversarial mechanism and build discriminative full-modal representation architecture using data of one/partial modalities. The main contributions are summarized as follows.

\begin{itemize}
	\item \textbf{Partial-modal representation} is proposed to deal with missing data modalities. Specifically, the partial-modal representation aims to obtain the transferable representation among data with different modalities. And when only partial-modal representations are accessible, the model can still generate a comprehensive description, as if constructed with data of all modalities.
	
	\item \textbf{Partial-modal GANs architecture} is proposed that can exploit the complementary benefits of all data channels with heterogeneous features using only one/partial channels. The generative model learns to fit the transferable distribution that characterizes the feature representation in the specific data channels that are likely to be missing in practice. Meanwhile, the discriminative model learns to judge whether the translated distribution is representative enough for the full modalities. Extensive experiment results reveal the effectiveness of the PM-GANs architecture, which outperform four state-of-the-art methods in the task of action recognition.
	
	\item \textbf{Partial-modal evaluation dataset} is newly introduced, which provides paired data of two different modalities--visible and infrared spectrum of human actions. Researchers can evaluate the transferable ability of the algorithms between the two modalities, as well as the discriminative ability of the generated representation by comparing with a series of baselines we provided in this paper. Meanwhile, the dataset can be used as a benchmark for bi-channel action recognition, since it is also carefully designed to serve for this purpose. The dataset contains more than 2,000 videos, 12 different actions, and to the best of our knowledge, is the first publicly available action recognition dataset that contains both infrared and visible spectrum.
\end{itemize}

The rest of the paper is organized as follows. In Section 2, we review the background and related works. In Section 3, we elaborate the details of our proposed method. Section 4 presents the newly-introduced dataset, its evaluations, and the experimental results of on it. Finally, Section 5 draws the conclusion.

\section{Related work}
\textbf{Transfer Learning and Cross-modal Representation}:
In the classical pattern recognition and machine learning tasks, sufficient training data that have variations in modality is clearly a desired but unrealistic goal \cite{shin2016deep,shao2015transfer}, thus restricting the representative ability of the model. Among the studies to address this problem, transfer learning attempts to transfer the feature space from a source domain to a target domain, and to lessen the adaption conflicts via domain adaption  \cite{ganin2015unsupervised,long2015learning,patel2015visual,tzeng2017adversarial}. The transferred knowledge type is not restricted to feature representation or instance, it also contains modality-correlation. With different aims, cross-dataset and cross-modal feature representation fall into feature-representation transfer by adapting the representations from different domains to a single common latent space, where features of multiple modalities are jointly learned and combined. Among these algorithms, Canonical Correlation Analysis (CCA) \cite{hardoon2004canonical,yeh2014heterogeneous} is a widely used one, which seeks to maximize the correlation between the projected vectors of two modalities. Another classical algorithm is Data Fusion Hashing (DFH) \cite{bronstein2010data} that embeds the input data from two arbitary spaces into a Hamming space in a supervised way. Differently, Cross-View Hashing (CVH) \cite{kumar2011learning} maximizes the weighted cumulative correlation and can be viewed as the general representation of CCA.

In recent years, with the renaissance of neural networks, many deep learning based transfer learning and cross-modal representation methods have been proposed as well. Bishifting Autoencoder Network \cite{kang2015learning} attempts to alleviate the discrepancy between the source and target datasets to the same space. To further take the feature alignment and auxiliary domain data into consideration, Aligned-to-generalized encoder (AGE) \cite{liu2018transferable} is proposed to map the aligned feature representations to the same generalized feature space with low intraclass variation and high interclass variation. Since GANs have been proposed by Goodfellow \cite{goodfellow2014generative} \emph{et al.} in 2014, a series of GANs-based methods have arisen for a wide variety of problems. Recently, a Cross-modal Generative Adversarial Networks for Common Representation Learning (CM-GANs) \cite{peng2017cm} is proposed. CM-GANs seeks to unify the inconsistent distribution and representation of different modalities by filling the heterogeneity of knowledge types like image and text. In contrast, we have completely different goal, which aims to use only partial data modalities to obtain a full-modal representation. Our focus is beyond the jointly-learned representation of multiple feature spaces, and takes one step further to achieve a discriminative partial-modality representation, which corresponds to our original aim of handling the problem of insufficient training data and data types.

\textbf{Infrared Action Recognition and Dataset}:
Most previous contributions \cite{rahmani2018learning,liu2017benchmarking} to the progress of action recognition have been made to the visible spectrum. 
Early approaches utilized the hand-crafted representation followed by classifiers, such as 3D Histogram of Gradient (HOG3D)   \cite{klaser2008spatio}, Histogram
of Optical Flow (HOF) \cite{laptev2008learning}, Space Time Interest Points
 (STIP) \cite{laptev2005space} and Trajectories \cite{wang2013dense}.
Wang et al. \cite{wang2016robust} proposed the Improved
Dense Trajectories (iDT) representation, making breakthroughs among  hand-crafted features.
In hand-crafted representation scheme, encoding methods such as Bag of Words (BoW) \cite{li2011contextual}, Fisher vector \cite{sanchez2013image}, VLAD \cite{delhumeau2013revisiting} are applied to aggregate the descriptors into video-level representation.
Benefiting from the success of  Convolutional Neural Networks (CNNs) in image classification, several deep network architectures have been proposed for action recognition.
Simonyan \emph{et al.} \cite{simonyan2014two} proposed a two-stream CNNs architecture which simultaneously captured appearance and motion information by  spatial and temporal nets.
Tran \emph{et al.} \cite{tran2015learning} investigated 3D ConvNets \cite{Ji20123D,karpathy2014large} in large-scale supervised training
datasets and effective deep architectures, achieving significant improvements.
Carreira \emph{et al.} \cite{carreira2017quo} designed a two-stream inflated 3D ConvNet, inflating filters and pooling kernels  into 3D to learn seamless spatio-temporal feature extractors.

Recently, increasing efforts have been devoted to infrared action recognition \cite{gao2016infar}. Corresponding to the classical methods employed in visible spectrum, spatiaotemporal representation for human action repretitive action recogtnition is also used under thermal imaging scenarios \cite{han2005human}. The combination of both visible and thermal imaging to improve human silhouette detection is also introduced by Han \emph{et al.} \cite{han2007fusion}. However, the scenario has not been studied where infrared data is available while the RGB channel is missing. The scenario has great potential in real-world of protecting privacy while benefiting the task of action recognition, and is meaningful to both the study of pattern recognition and the welfare of the community at large. Therefore, we are motivated to dedicate to improving the situation by constructing a robust and discriminative partial-modal representations, and to specify action recognition as the case in this paper.

There is no publicly available infrared action recognition dataset except the infrared action recognition dataset (InfAR)\cite{gao2016infar}. To the best of our knowledge, there remains no publicly available action recognition dataset that contains both infrared and visible videos. In this paper we introduce a new dataset that provides paired data of infrared and visible spectrum. The dataset contains large variety in action classes and samples, taking multiple aims into consideration. Researchers can evaluate their methods on visible, thermal data channels, or to use them combined. More importantly, it can be used to evaluate the transferable ability between the two data modalities, and the discriminative ability of the jointly-learned representation. Our dataset will be made as part of the submission and benefit the public.

\section{Proposed approach}
\label{sec:actiongan}

The overall pipeline of the proposed PM-GANs for action recognition is shown in Fig. \ref{fig:framework}. Our goal is to generate a full-modal representation using only the partial modalities. The framework learns the transferable representation among different data channels based on a conditional generator networks. Based on the transferred representation, the framework build a discriminative full-modal representation network using only part of the data channels.


\subsection{Transferable basis for partial modality}

The transferable ability with the PM-GANs architecture is the basis for the construction of full-modal representation with partial modality. We assume that there exists a mapping from an observed distribution $f_{Vis}$ and an input distribution $f_{Inf}$, producing an output representation which shares the feature of the observed $f_{Vis}$. Therefore, we attempt to learn a generator $G$ to generate the feature distribution of the missing data channel $f_{Vis}$ from the partially available distribution denoted as $f_{Inf}$.
Based on the scheme of conditional generator networks, the generator immediately transform the partially available distribution $f_{Inf}$ and noise $z$ to output the missing distribution via the following equation:  
\begin{equation}
\begin{aligned}
\min\limits_G\max\limits_D
\mathcal{L}(G,D)=&\mathbb{E}_{f_{Vis}\thicksim P_{data}(f_{Vis})}[\log D(f_{Vis})]+\\
&\mathbb{E}_{f_{Inf}\thicksim P_{data}(f_{Inf}),z\thicksim P_{data}(z)}[\log(1-D(G(f_{Inf},z))) ],
\end{aligned}
\end{equation}
where $G(f_{Inf},z)$ denotes the output distribution. 
The input distribution $f_{Inf}$ and observed distribution $f_{Vis}$ denote the data of infrared and RGB channels respectively in our action recognition task. 
The generator $G$ is designed to minimize this objective to fake the generated distribution as well as possible, while the the real output feature discriminator $D$ tries to maximize its accuracy of telling the real from the fake one.

In this work, the discriminator is also designed for pattern recognition.
Thus, another prediction loss is explored:
\begin{equation}
\begin{aligned}
\mathcal{L}_p(G,D_p)=\mathbb{E}_{f_{Inf}\thicksim P_{data}(f_{Inf}),z\thicksim P_{data}(z)}[L_{cls}(l,D_p(f_{Inf},G(f_{Inf},z)))],
\end{aligned}
\end{equation}
where $l$ denotes the correct label of partially available data samples, in the form of one-hot vector, and $L_{cls}$ is log loss over the predicted class confidences vector and the ground truth label.
For convenience, we denote the discriminator part and the predictor part of discriminative net as $D_d$ and $D_p$ respectively.
Finally, the objective function can be formulated as:
\begin{equation}
\begin{aligned}
\min\limits_G\max\limits_D
\mathcal{L}(G,D)&=\mathbb{E}_{f_{Vis}\thicksim P_{data}(f_{Vis})}[\log D(f_{Vis})]\\&+\mathbb{E}_{f_{Inf}\thicksim P_{data}(f_{Inf}),z\thicksim P_{data}(z)}[\log(1-D_d(G(f_{Inf},z))) ]\\
&+\mathbb{E}_{f_{Inf}\thicksim P_{data}(f_{Inf}),z\thicksim P_{data}(z)}[-L_{cls}(l,D_p(f_{Inf},G(f_{Inf},z)))].
\end{aligned}
\end{equation}

\subsection{Transferable Net}

\begin{figure}[htbp]
	\centering
	\includegraphics[width=10cm]{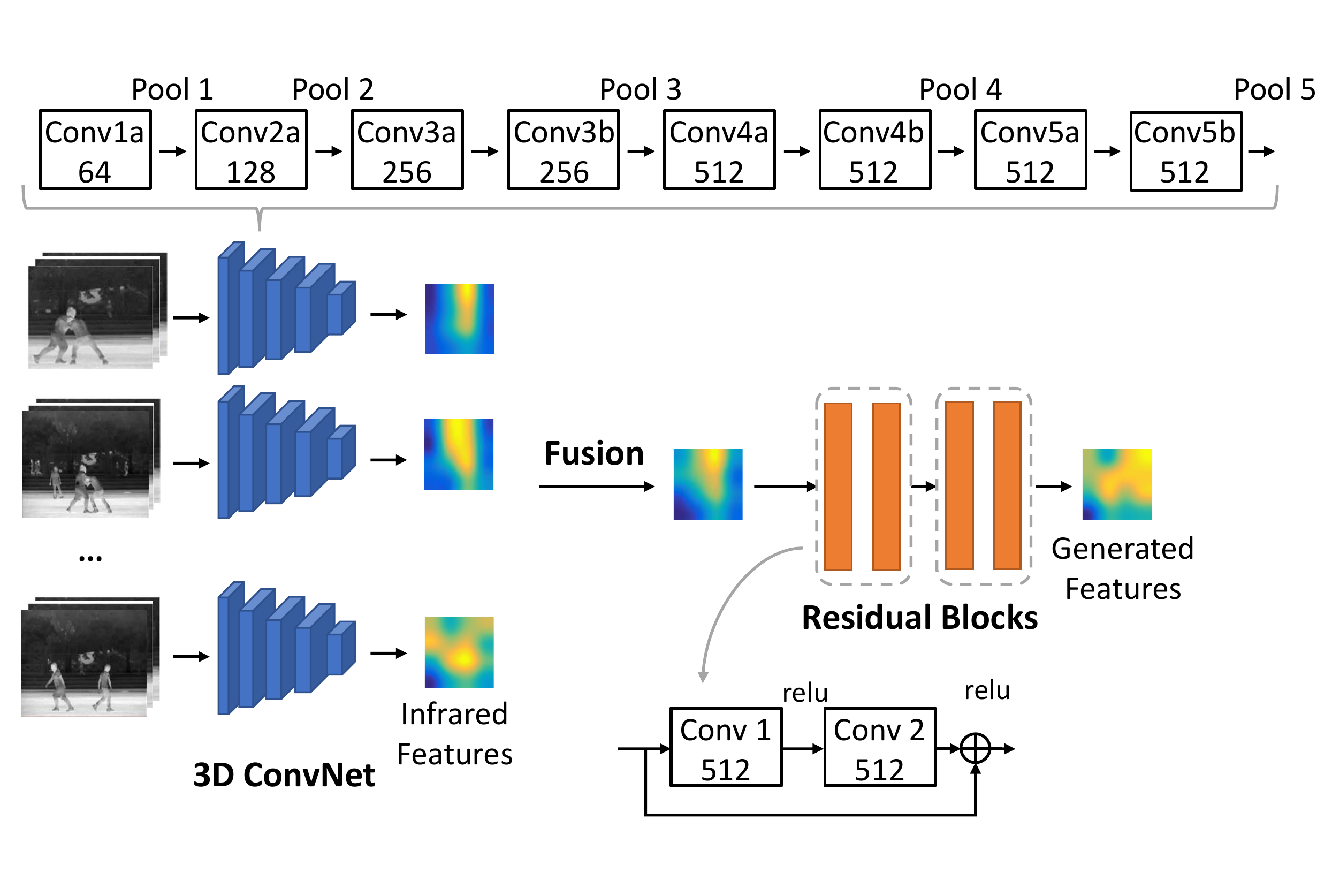}
	\vspace{-2ex}
	\caption{The proposed transferable generative net is built upon the C3D network \cite{tran2015learning}. Video clips are sent to 3D ConvNet to obtain feature maps of each clip $f_{Inf}^{(i)}$ and all feature maps are fused as $f_{Inf}$ to represent the whole action video. Then, residual blocks are added to this net to produce fake feature maps $f_g$ similar to the visible spectrum. }
	\label{fig:generativenet}
	\vspace{-0.6cm}
\end{figure}

The transferable net simulates the target distribution from the convolutional feature map of the partially-available data distribution, which, as shown in Fig.  \ref{fig:generativenet}, are then made as an input of the generator to obtain feature maps of missing distribution. For each input distribution, denoted as $\{f_{Inf}^{(1)},f_{Inf}^{(2)},\dots,f_{Inf}^{(T)}\}$, where $f_{Inf}^{(i)} \in \mathbb{R}^{H \times W \times D} $ and $H,W,D$ denote the width, height and number of channels of feature maps.
To incorporate all feature maps into a high-level representation, the sum fusion model in \cite{feichtenhofer2016convolutional} is applied to compute the sum of $T$ feature maps at the same spatial location $i, j$ and feature channel $d$:
\begin{equation}
f_{Inf}^{sum}(i,j,d)=f_{Inf}^{(1)}(i,j,d)+f_{Inf}^{(2)}(i,j,d)+\dots+f_{Inf}^{(T)}(i,j,d),
\end{equation}
where $1 \leq i \leq H$, $1 \leq j \leq W$, $1 \leq d \leq D$.
The final feature map of the input distribution is computed as the average value of sum feature map $f_{Inf}^{sum}$ in each location, denoted as $f_{Inf}$.
Then the generator takes the final input feature map and generate the fake target feature map, $G(f_{Inf},z)$.
The generator consists of two residual blocks \cite{he2016deep} to produce feature map with the same size as infrared feature map.
Thus, the generative loss $L_G$ is expressed as:
\begin{equation}
L_G=-\log(D_d(G(f_{Inf},z)).
\end{equation}


\subsection{Discriminative net using partial modality}

To enable the generative net to produce full-modal representation which incorporates the complementary benefits among data of different modalities, 
a two-part discriminative net is designed, as shown in Fig. \ref{fig:framework}.
The discriminative net contains a discriminator part and a predictor part.
The discriminator part follows the scheme of conventional  discriminator in GAN which is applied to distinguish between real and fake visible feature map in order to boost the quality of generated fake feature.
Specifically, the discriminator part consists of a fully-connected layer followed by a sigmoid function,
which produces an adversarial loss.
Thus, the adversarial loss $L_a$ is defined as:
\begin{equation}
L_a=-\log D_d(f_{Vis})-\log(1-D_d(G(f_{Inf,z}))),
\end{equation}
where $L_a$ encourages the discriminator network to distinguish the  generated target feature representation from real one.

The predictor aims to boost the accuracy of assigning the right label to each feature distribution.
It consists of a fully-connected layer followed by a softmax layer which takes the fusion of the feature map of both the partially-available data channel and generated missing channel and finally outputs the category-level confidences.
To fuse these two feature maps, a convolutional fusion model in \cite{feichtenhofer2016convolutional} is applied to automatically learn the fusion weights:
\begin{equation}
f_{conv}=f_{cat}*\textbf{f}+b,
\end{equation}
where \textbf{f} are filters with dimensions $1 \times 1 \times 2D \times D$, and $f_{cat}$ denotes the stack of two feature maps at the same spatial locations $(i,j)$ across the feature channels $d$:

\begin{equation}
\begin{aligned}
f_{cat}^{(i,j,2d)}&=f_{Inf}^{(i,j,d)},\\  f_{cat}^{(i,j,2d-1)}&=f_g^{(i,j,d)},
\end{aligned}
\end{equation}
where $f_g$ denotes the  generated fake feature map $G(f_{Inf},z)$.

Thus, the predictive loss $L_p$ can be formulated as:
\begin{equation}
L_p=-\log l \cdot D_p(f_{Inf},G(f_{Inf},z)).
\end{equation}
Thus, the final discriminative loss $L_D$ can be defined as the weighted sum of adversarial loss and predictive loss:
\begin{equation}
L_D= w_1 \cdot L_a + w_2 \cdot L_p.
\end{equation}

In the training process, the transferable net and the full-modal discriminative net are alternatively trained until the generated feature of missing channel becomes close to real and the discriminative net achieves precise recognition.
Detailed training strategies will be given in Section \ref{seq:experi}.
In the testing process, we only need to send one/part of the data modality into the PM-GANs framework, and the generative net will automatically generate a transferred feature representation for the missing modality, and 
the predictor of discriminative net construct a full-modal representation and predict the label.

\section{Experiments}
\label{seq:experi}
In this section, we first introduce our new dataset for partial-modality infrared action recognition. In detail, the specifications and a complete evaluation of the dataset will be elaborated. For the experiment part, we introduce the configurations of the experiments and show the results and analyses corresponding to our method. Specifically our experiments are in three folds. First, we assess the effectiveness of the transferable net by comparing the generated feature representations with the real ones. Second, we evaluate the discriminative net ability constructed using partial data modality. Finally, we compare our approach with four state-of-the-art methods to verify the effectiveness of the PM-GANs.

\subsection{Cross-modal Infrared-Visible Dataset for Action Recognition}
\begin{figure}[htbp]
	\centering
	\vspace{-2ex}
	\includegraphics[width=11cm ]{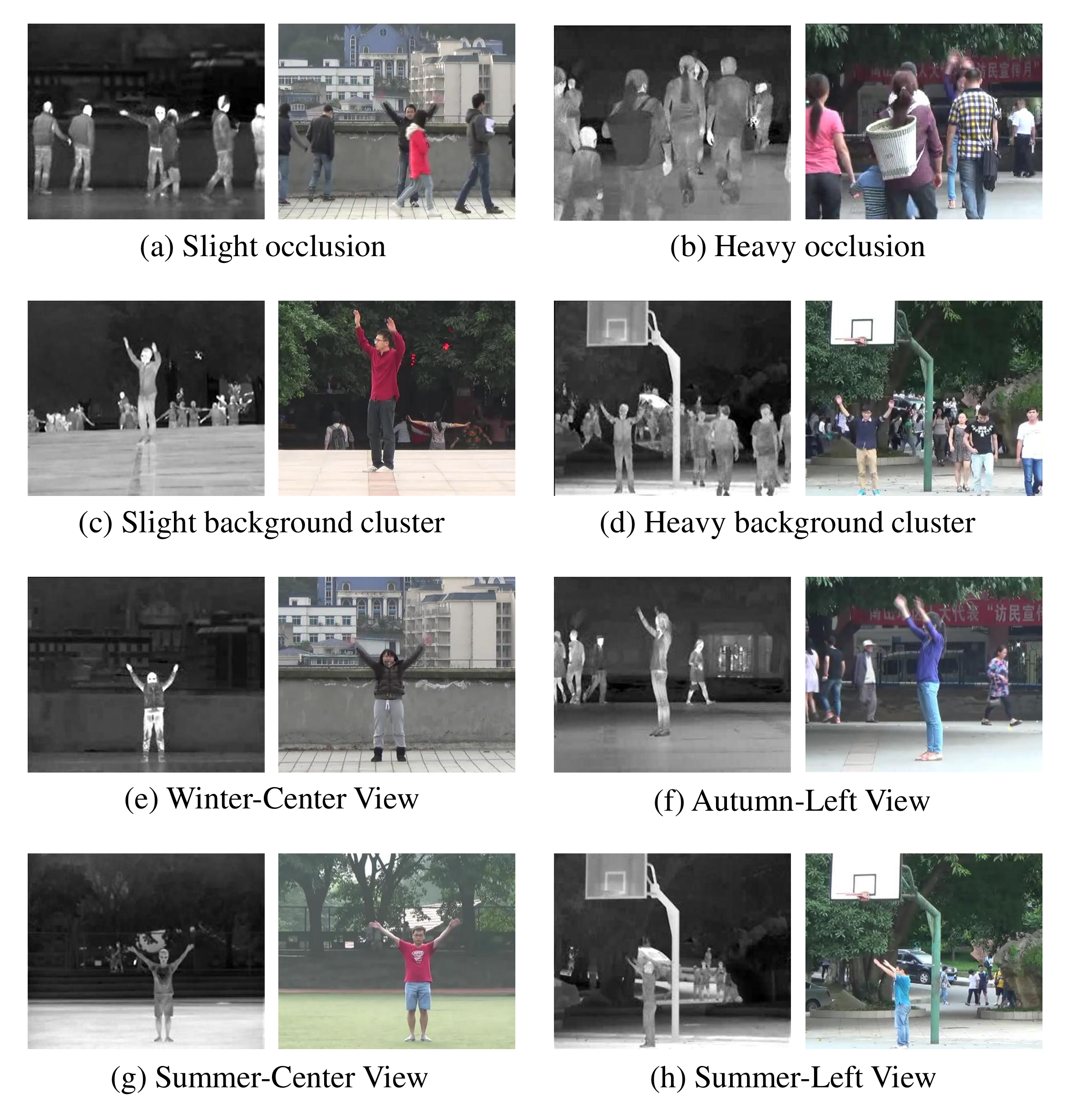}
	\caption{Example paired frames for the action ``wave2" in the newly introduced multi-modal dataset for action recognition.
		The left ones are in infrared channel and the right ones are in RGB channel.}
	\label{fig:examples}
	\vspace{-0.6cm}
\end{figure}

We introduce a new action recognition dataset, which is constructed by paired videos of both RGB and infrared data channels. Each action class contains a singular action type, and each video sample contains one action class. In total there are 12 classes of both individual action and person-person interactive actions. For individual actions: 
one hand wave (wave1), multiple hands wave  (wave2), handclap, walk, jog, jump, skip, and interactive actions: handshake, hug, push, punch and fight. For each action class, there are 100 paired videos, with a frame rate of $25$ fps. 
The frame resolutions are $293 \times 256$ for infrared channel and  $480\times720$ for RGB channel. 
Each actions is performed by 50 different volunteers.
A sample of the frames are illustrated in Fig. \ref{fig:examples}.  
The duration distribution of the dataset is listed in Table \ref{tab:duration}.


In order to simulate the real-world variations, four scenario variables are considered: the background complexity, season, occlusion, and viewpoint. 
\\{\bfseries Background Complexity:} 
In our newly-introduced dataset, the 
background varies from relatively simple scenes (plain background) to
complex ones (with moving objects). For simple
background, there are only one or two people performing actions, as
shown in Fig.\ref{fig:examples} (c). While for complex background, interrupting pedestrian activities concur with the objective action in different degrees, as shown
in Fig. \ref{fig:examples} (d). 
\\{\bfseries Season:} 
The infrared channel is heavily effected by the seasons, because it reflects the heat radiation of objects.
In winter, when ambient temperature is in a low value, 
the imaging of human body is salient and clear.
However, in summer, the contrast between human and background is ambiguous.
Thus, we divide the seasons into three categories: winter, spring/autumn, summer, as shown in Fig. \ref{fig:examples} (e)-(h).
The video number proportions of these three seasons are $30\%$, $50\%$, and $20\%$, respectively.
All actions were performed in these three seasons.
\\{\bfseries Occlusion:} 
Specific videos with
occlusions from 0\% to over 50\% are arranged in each action class to promote the diversity and complexity of dataset, as shown in Fig. \ref{fig:examples} (a)-(b).
\\{\bfseries Viewpoint:} 
The variation of different viewpoints is also an important factor considered.
The video clips under the front-view, left-side-view, right-side-view are all included in the dataset, as shown in Fig. \ref{fig:examples} (e)-(h).

\begin{table}[htbp]
	\centering
	\vspace{-3ex}
	\caption{Duration  distribution of videos per class.}
	\begin{tabular}{lllllll}
		\toprule
		\multicolumn{1}{l}{\textbf{Action}} &       & \multicolumn{5}{l}{\textbf{Duration}} \\
		\cmidrule{3-7}          &       & \textbf{0-5s} &       & \textbf{5-10s} &       & \textbf{\textgreater 10s} \\
		\multicolumn{1}{l}{Fight} &       & 27    &       & 66    &       & 7 \\
		\multicolumn{1}{l}{Handclapping} &       & 45    &       & 53    &       & 2 \\
		\multicolumn{1}{l}{Handshake} &       & 35    &       & 61    &       & 4 \\
		\multicolumn{1}{l}{Hug} &       & 26    &       & 72    &       & 2 \\
		\multicolumn{1}{l}{Jog} &       & 88    &       & 12    &       & 0 \\
		\multicolumn{1}{l}{Jump} &       & 73    &       & 27    &       & 0 \\
		\multicolumn{1}{l}{Punch} &       & 55    &       & 45    &       & 0 \\
		\multicolumn{1}{l}{Push} &       & 55    &       & 45    &       & 0 \\
		\multicolumn{1}{l}{Skip} &       & 82    &       & 18    &       & 0 \\
		\multicolumn{1}{l}{Walk} &       & 63    &       & 37    &       & 0 \\
		\multicolumn{1}{l}{Wave1} &       & 40    &       & 59    &       & 1 \\
		\multicolumn{1}{l}{Wave2} &       & 42    &       & 58    &       & 0 \\
		\bottomrule
		\vspace{-1cm}
	\end{tabular}%
	\label{tab:duration}%
\end{table}%

We split $75$\% of the paired video clips couples as the training set, and the rest as the testing set.

To investigate the suitable representations for each spectrum and the most complimentary representation couples, we select several effective representations to test their discriminative ability on both RGB and infared channels, and the combined channels.

We feed the original video clips , the MHI image clips \cite{Bobick2001The} and the optical flow clips \cite{laptev2008learning}, denoted as ``Org", ``MHI", ``Optical Flow" into the 3D-CNN \cite{tran2015learning} to obtain spatiotemporal features.
The 3D-CNN  takes a 16-frame clip as inputs and performs 3D convolution and 3D pooling, which calculate the appearance and motion information simultaneously.
Specifically, we extract the output of the last fully connected layer and conduct a max pooling to all clip features of one video.
In the case of two-modality fusion, we directly concatenate the features of infrared channel and RGB channel.
After that, a linear SVM classifier is trained to obtain the final recognition results.
The 3D-CNN is fine-tuned by the corresponding  maps of our training set. 


\begin{table}[htbp]
	\centering
	\vspace{-3ex}
	\caption{The evaluation results of different features on different channels and their fusion on the proposed dataset.}
	\begin{tabular}{llllr}
		\toprule
		\textbf{Method} &       & \textbf{Descriptor} &       & \textbf{Accuracy(\%)} \\
		\midrule
		\textbf{Infrared Channel} &       & Org   &       & 55\% \\
		&       & Optical Flow &       & 69.67\% \\
		&       & MHI   &       & 61\% \\
		\textbf{RGB Channel} &       & Org   &       &  49\% \\
		&       & Optical Flow &       & 78.66\% \\
		&       & MHI   &       &  65.33\%\\
		\textbf{Fusion} &       & Org &       &  55.33\% \\
		&       & Optical Flow &       & 80.67\% \\
		
		&       & MHI &       &  68.67\% \\
		
		\bottomrule
		\vspace{-1cm}
	\end{tabular}%
	\label{tab:features}%
\end{table}%

As shown in Table \ref{tab:features}, the performance of different representations for both infrared and RGB channels and their combined results are listed.
It is clearly observed that for both infrared and RGB channel, the 3D-CNN features after optical flows achieve the best performance.
In two modalities fusion, the 3D-CNN features after optical flows in RGB channel can effectively boost the performance of using the infrared channel only.
Thus, in the following experiments of transferable nets and discriminative nets \cite{tzeng2017adversarial}, the optical flows are selected as input for representation learning via PM-GANs.



\subsection{Implementation Details}
For the input data, we compute optical flows using the toolbox of Liu \cite{liu2009beyond}.
The 3D ConvNet in transferable generative net is fine-tuned on the infrared optical flows of training set.
And the adversarial visible feature maps are extracted from a 3D ConvNet fine-tuned on the visible optical flows of training set.
The sampled numbers of clips $T$ is set as 5, and each clip has a duration of 16 frames.
The loss weights $w_1$ and $w_2$ are set as $0.1$ and $0.9$ respectively.
 We set the initial learning rate at $2 \times 10^{-5}$.
The whole network is trained with ADAM optimization algorithm \cite{kingma2014adam}
with $\beta_1 = 0.5$ and $\beta_2 = 0.999$, batch size of 30 on a single NVIDIA GeForce GTX TITAN X GPU with 12GB memory.  The framework is implemented using TensorFlow library and accelerated by CUDA 8.0.

\begin{figure}[htbp]
	\centering
	\vspace{-3ex}
	\includegraphics[width=10cm ]{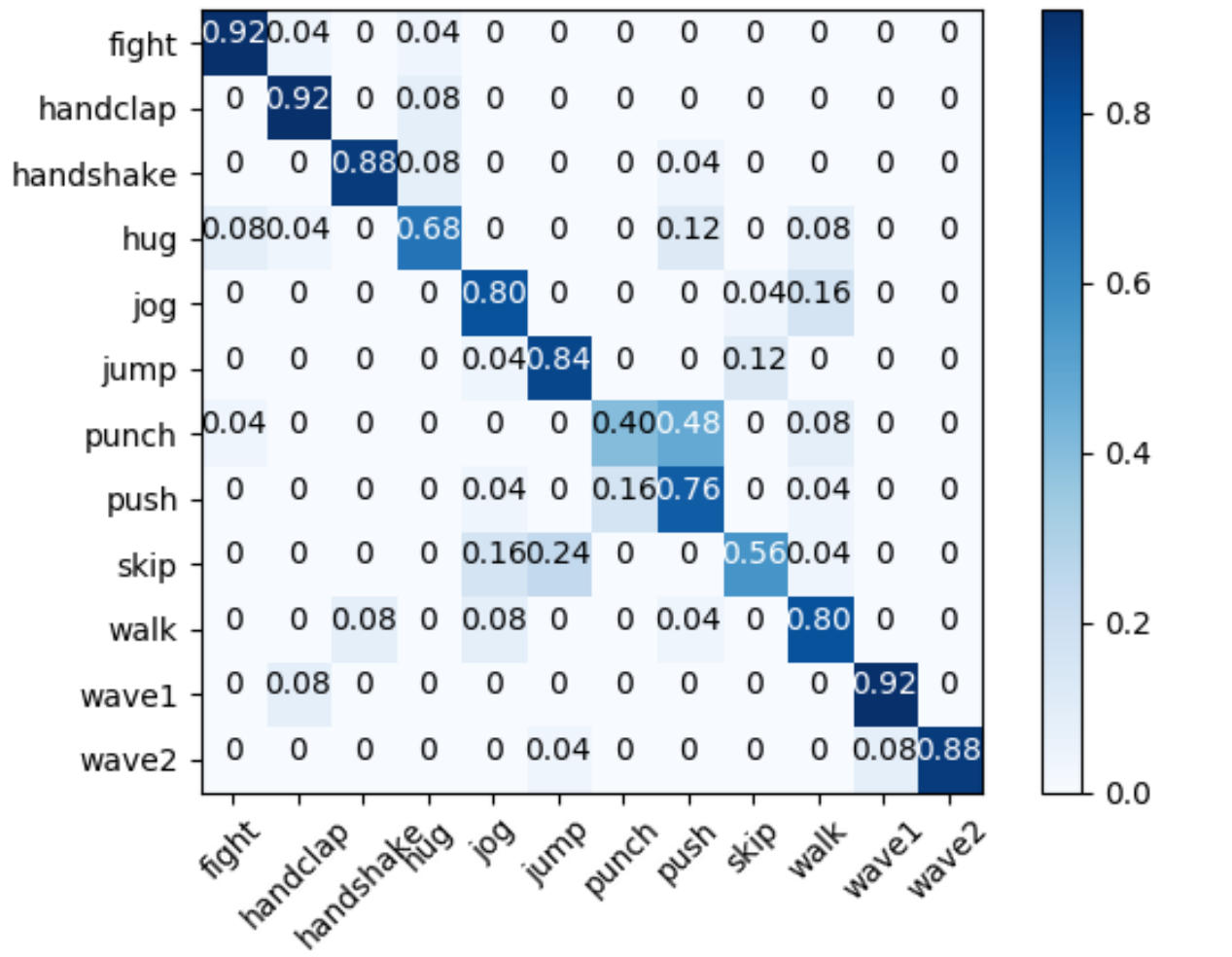}
	\caption{The results illustrated in confusion matrices using the proposed method.}
	\label{fig:mat}
	\vspace{-1cm}
\end{figure}

\begin{table}[htbp]
	\centering
	\vspace{-3ex}
	\caption{Evaluation results on the discriminative ability of transferable modality.}
	\begin{tabular}{llr}
		\toprule
		\textbf{Data Modalities} &       & \textbf{Accuracy (\%)} \\
		\midrule
		Infrared channel&       & 71.67\% \\
		RGB channel   &       & 79.33\% \\
		Generated RGB  &       & 76.67\% \\
		Infared + RGB channels &       & 82.33\% \\
		Infrared channel + Generated RGB     
		&       & 78\% \\
		\bottomrule
		\vspace{-1cm}
	\end{tabular}%
	\label{tab:effectiveness}%
\end{table}%

\subsection{Transferable Net Evaluations}

The PM-GANs model is evaluated on the proposed action recognition datatset.
We present the results of five different modalities as shown in Table \ref{tab:effectiveness}.
For single modality, we utilize the 3D ConvNet part and the predictor part without fusion model for training and testing.
And for the case of real infrared and RGB channel fusion, we directly input the real feature map of RGB channel to the fusion model instead of using generated ones.
From Table \ref{tab:effectiveness}, we can observe the generated RGB representations perform better than the original infrared ones,  which shows that the PM-GANs have indeed discovered useful information through modality transfer.
Moreover, the fusion of infrared and generated RGB representations achieves an Accuracy of 78\%.
Although it performs worse than the original RGB channels and the fusion of infrared and  RGB channels, it only utilizes the information of infrared channel in the testing process.

In order to analyze the intra-class performance, the
confusion matrices are drawn in Fig. \ref{fig:mat}.
As observed, the proposed method generally shows
good performance in action classification: in most classes, the testing samples are
assigned the correct label.
However, we notice that the ``punch", ''skip" action samples are likely to be classified as ``push" and ``jump" respectively.
One likely reason is that two sets of actions are similar in both movement and process, sometimes even hard to distinguish for human eyes.

To get insight into how effective the transferable ability of PM-GANs are, we rearrange the training and testing splits.
Specifically, we utilize the scenes of Spring/Autumn and Summer for training, and Winter for testing.
We use this split to examine the generalization ability of the proposed model.
As can be seen in Table \ref{tab:2ressplit},
the generated fake RGB representations outperform the original infrared ones, which shows the robust transferability of PM-GANs.

\begin{table}[htbp]
  \centering
  \vspace{-3ex}
  \caption{Evaluation of the models generalization ability using a separate dataset.}
    \begin{tabular}{llr}
    \toprule
    \textbf{Modalities of a seperate dataset} &       & \textbf{Accuracy (\%)} \\
    \midrule
    Infrared channel &       & 74.17\% \\
    RGB channel   &       & 79.44\% \\
    generated RGB &       & 77.78\% \\
    Infared+RGB channels &       & 82.78\% \\
    Infrared channel+ Generated RGB &       & 80.28\% \\
    \bottomrule
    \vspace{-1 cm}
    \end{tabular}%
  \label{tab:2ressplit}%
\end{table}%

\subsection{Comparisons with Other Methods}

To evaluate the effectiveness of PM-GANs, we compared  our method with four state-of-the-art methods, including the most effective handcrafted features iDT \cite{wang2016robust}, and  the state-of-the-art deep architecture \cite{tran2015learning}.
In addition, we also compare our methods with two state-of-the-art framework for infrared action recognition \cite{gao2016infar,Jiang2017Learning}. 
For iDT features,  Fisher Vector \cite{perronnin2010improving} is applied to encode and then a linear SVM classifier \cite{schuldt2004recognizing} is trained for action classification.
As for the C3D architecture, the network is fine-tuned by the proposed training dataset.
Then max popling followed by a SVM classifier is applied as the evaluation in Table \ref{tab:features}.
For \cite{gao2016infar}, we followed the original experimental settings provided by the author.
For \cite{Jiang2017Learning}, we implement and select the configuration with the optimal results based on the original submission.
We apply the discriminative code layer and the second fusion strategy for feature extraction, and train a K-nearest neighbor classifier (KNN) \cite{bui2017novel} using the provided Gaussian kernel function for classification.
Note that all of the results are achieved using  unified optical flows as the input.

\begin{table}[htbp]
	\centering
	\vspace{-3ex}
	\caption{Comparisons with four state-of-the-art approaches}
	\begin{tabular}{lr}
		\toprule
		\textbf{Method} & \textbf{Accuracy (\%)} \\
		\midrule
		iDT \cite{wang2016robust}   & 72.33\% \\
		C3D\ \cite{tran2015learning}   & 69.67\% \\
		Two-Stream CNN \cite{gao2016infar} & 68\% \\
		Two-Stream 3D-CNN \cite{Jiang2017Learning} & 74.67\% \\
		PM-GANs & 78\% \\
		\bottomrule
	\vspace{-1cm}
	\end{tabular}%
	\label{tab:comparisons}%
\end{table}%

Table \ref{tab:comparisons} presents the accuracy of the competing approaches.
As observed, the handcrafted iDT method achieves comparable results with some high-level architecture.
Methods using 3D-CNN outperform the method with 2D-CNN architecture.
One reason to explain is that the 3D-CNN architecture is better in modeling temporal variations.
The two-stream 3D-CNNs outperform the conventional iDT framework and robust C3D models, showing effective strength of the proposed discriminative code layer.
Our proposed PM-GANs achieves the highest accuracy, which shows the effectiveness of the transferred feature representation and the robustness of our constructed model using only part of the data modalities.

\section{Conclusions}

In this paper, we proprosed a novel Partial-modal Generative Adversarial Networks (PM-GANs) to construct a discriminative full-modal representation with only part of the data modalities being available. Our methods learns the transferable representation among hetergeneous data modalities using adversarial learning, and build a discriminative net that represents all modalities. Our method is evaluated in the task of action recogntion and outperforms four state-of-the-art methods on four newly-introduced dataset. The dataset, which contains paired videos in both infrared and visible spectrum, will be made as the first publicly available visible-infrared dataset for action recognition.


\clearpage

\bibliographystyle{splncs}
\bibliography{egbib}
\end{document}